\pdfoutput=1

\documentclass[11pt]{article}

\usepackage{EMNLP2022}

\usepackage{times}
\usepackage{latexsym}

\usepackage[T1]{fontenc}

\usepackage[utf8]{inputenc}

\usepackage{microtype}

\usepackage{inconsolata}

\usepackage{caption}
\usepackage{graphicx}
\usepackage{subcaption}
\usepackage{multirow}
\usepackage[normalem]{ulem}
\useunder{\uline}{\ul}{}
\usepackage{amsmath}
\usepackage{amssymb}
\usepackage{mathtools}
\usepackage{amsthm}
\usepackage{url}

\usepackage[capitalize,noabbrev]{cleveref}

\theoremstyle{plain}
\newtheorem{theorem}{Theorem}[section]
\newtheorem{proposition}[theorem]{Proposition}

\theoremstyle{definition}

\theoremstyle{remark}

\title{
    Textual Manifold-based Defense Against \\
    Natural Language Adversarial Examples
}

\author{Dang Minh Nguyen \\
  VinAI Research, Vietnam \\
  \texttt{v.dangnm12@vinai.io} \\\And
  Luu Anh Tuan\thanks{~~Corresponding author} \\
  Nanyang Technological University, Singapore \\
\texttt{anhtuan.luu@ntu.edu.sg} \\}

\begin{document}
\maketitle
\begin{abstract}
Recent studies on adversarial images have shown that they tend to leave the underlying low-dimensional data manifold, making them significantly more challenging for current models to make correct predictions. This so-called off-manifold conjecture has inspired a novel line of defenses against adversarial attacks on images. In this study, we find a similar phenomenon occurs in the contextualized embedding space induced by pretrained language models, in which adversarial texts tend to have their embeddings diverge from the manifold of natural ones. Based on this finding, we propose Textual Manifold-based Defense (TMD), a defense mechanism that projects text embeddings onto an approximated embedding manifold before classification. It reduces the complexity of potential adversarial examples, which ultimately enhances the robustness of the protected model. Through extensive experiments, our method consistently and significantly outperforms previous defenses under various attack settings without trading off clean accuracy. To the best of our knowledge, this is the first NLP defense that leverages the manifold structure against adversarial attacks. Our code is available at \url{https://github.com/dangne/tmd}.
\end{abstract}

\section{Introduction}
\label{introduction}
The field of NLP has achieved remarkable success in recent years, thanks to the development of large pretrained language models (PLMs). However, multiple studies have shown that these models are vulnerable to adversarial examples - carefully optimized inputs that cause erroneous predictions while remaining imperceptible to humans \cite{Jin_Jin_Zhou_Szolovits_2020, li-etal-2020-bert-attack, garg-ramakrishnan-2020-bae}. This problem raises serious security concerns as PLMs are widely deployed in many modern NLP applications. 

Various defenses have been proposed to counter adversarial attacks, which can be broadly categorized into empirical \cite{si-etal-2021-better,dong2021should,DBLP:conf/iclr/MiyatoDG17} and certified \cite{jia-etal-2019-certified,ye-etal-2020-safer,DBLP:journals/corr/abs-2105-03743,zhao2022certified} defenses. Adversarial training is currently the most successful empirical method \cite{pmlr-v80-athalye18a, DBLP:conf/iclr/DongLJ021,zhou-etal-2021-defense}. It operates by jointly training the victim model on clean and adversarial examples to improve the robustness. However, one major drawback of this approach is the prohibitively expensive computational cost. \citet{NEURIPS2018_f708f064} has theoretically shown that with just simple models, the sample complexity of adversarial training already grows substantially compared to standard training. Alternatively, certified defenses aim to achieve a theoretical robustness guarantee for victim models under specific adversarial settings. However, most certified defenses for NLP are based on strong assumptions on the network architecture \cite{jia-etal-2019-certified,DBLP:conf/iclr/ShiZCHH20}, and the synonym set used by attackers is often assumed to be accessible to the defenders \cite{ye-etal-2020-safer}. \citet{li-etal-2021-searching} has shown that when using different synonym set during the attack, the effectiveness of these methods can drop significantly.

Concurrent with the streams of attack and defense research, numerous efforts have been made to understand the characteristics of adversarial examples \cite{DBLP:journals/corr/SzegedyZSBEGF13,DBLP:journals/corr/GoodfellowSS14, https://doi.org/10.48550/arxiv.1608.07690, DBLP:conf/iclr/GilmerMFSRWG18, DBLP:conf/nips/IlyasSTETM19,DBLP:conf/iclr/TsiprasSETM19}. One rising hypothesis is the off-manifold conjecture, which states that adversarial examples leave the underlying low-dimensional manifold of natural data \cite{https://doi.org/10.48550/arxiv.1608.07690, DBLP:conf/iclr/GilmerMFSRWG18,Stutz_2019_CVPR,DBLP:journals/corr/abs-2106-10151}. This observation has inspired a new line of defenses that leverage the data manifold to defend against adversarial examples, namely manifold-based defenses \cite{DBLP:conf/iclr/SamangoueiKC18,DBLP:conf/ccs/MengC17,DBLP:journals/corr/abs-1710-10766}. Despite the early signs of success, such methods have only focused on images. It remains unclear if the off-manifold conjecture also generalizes to other data domains such as texts and how one can utilize this property to improve models' robustness.

In this study, we empirically show that the off-manifold conjecture indeed holds in the contextualized embedding space of textual data. Based on this finding, we develop Textual Manifold-based Defense (TMD), a novel method that leverages the manifold of text embeddings to improve NLP robustness. Our approach consists of two key steps: (1) approximating the contextualized embedding manifold by training a generative model on the continuous representations of natural texts, and (2) given an unseen input at inference, we first extract its embedding, then use a sampling-based reconstruction method to project the embedding onto the learned manifold before performing standard classification. TMD has several benefits compared to previous defenses: our method improves robustness without heavily compromising the clean accuracy, and our method is structure-free, i.e., it can be easily adapted to different model architectures. The results of extensive experiments under diverse adversarial settings show that our method consistently outperforms previous defenses by a large margin. 

In summary, the key contributions in this paper are as follows:
\begin{itemize}
    \item We show empirical evidence that the off-manifold conjecture holds in the contextualized embedding space induced by PLMs.
    \item We propose TMD, a novel manifold-based defense that utilizes the off-manifold conjecture against textual adversarial examples.
    \item We perform extensive experiments under various settings, and the results show that our method consistently outperforms previous defenses.
\end{itemize}
\begin{figure*}
\begin{center}
\centerline{\includegraphics[width=\textwidth]{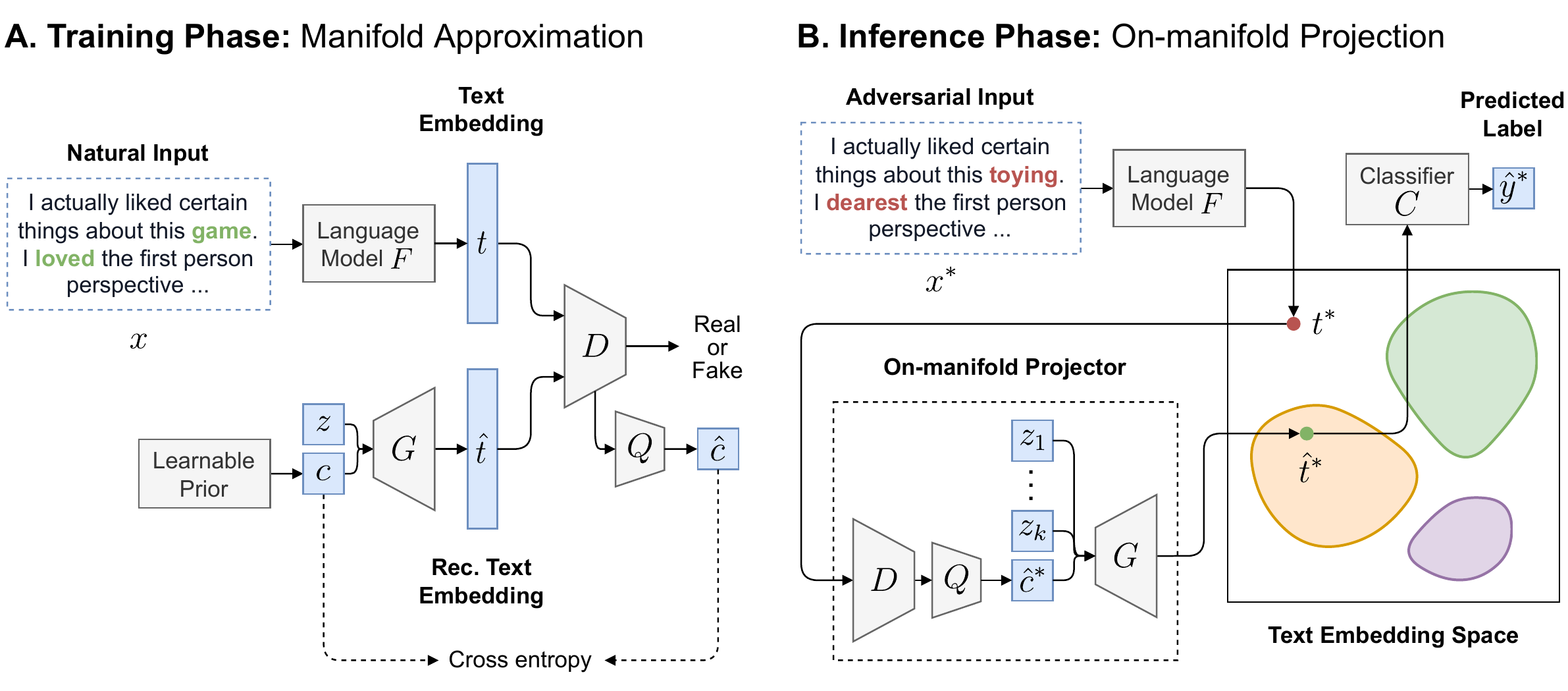}}
 \vspace{1mm}
\caption{An overview of the Textual Manifold-based Defense. (A) Training Phase: All textual inputs $x$ are transformed into continuous representations $t = F(x)$. An InfoGAN \cite{DBLP:conf/nips/ChenCDHSSA16} with learnable prior is trained to distinguish between real embeddings $t$ versus fake embeddings $\hat{t}$ to implicitly learn the natural text embedding manifold. (B) On-manifold projection: Once the generative model is trained, novel input embedding $t^* = F(x^*)$ is projected onto the approximated manifold using a sampling-based reconstruction strategy. The reconstructed embedding $\hat{t}^*$ is fed to the classifier $C(.)$ to produce the final predicted label $\hat{y}^*$. The colored blobs represent the approximated disjoint submanifolds in the contextualized embedding space.}
\vspace{-2.5mm}
\label{fig:overview}
\end{center}
\end{figure*}

\section{Related Work}
\label{related_work}
\subsection{Adversarial Attacks and Defenses}
Textual adversarial examples can be generated at different granularity levels. One can add, delete, replace characters \cite{JiDeepWordBug18,ebrahimi2018hotflip,DBLP:conf/ndss/LiJDLW19} or words \cite{ren-etal-2019-generating,Jin_Jin_Zhou_Szolovits_2020,li-etal-2020-bert-attack,garg-ramakrishnan-2020-bae,alzantot-etal-2018-generating}, or manipulate the entire sentence \cite{iyyer-etal-2018-adversarial,ribeiro-etal-2018-semantically,DBLP:conf/iclr/ZhaoDS18} to maximize the prediction error without changing the original semantics. Among the different attack strategies above, word substitution-based attacks are the most popular and well-studied methods in the literature \cite{ebrahimi-etal-2018-hotflip, alzantot-etal-2018-generating, ren-etal-2019-generating, Jin_Jin_Zhou_Szolovits_2020, li-etal-2020-bert-attack}. \citet{ebrahimi-etal-2018-hotflip} is the first work to propose a white box gradient-based attack on textual data. Follow-up works \cite{alzantot-etal-2018-generating, ren-etal-2019-generating, Jin_Jin_Zhou_Szolovits_2020, li-etal-2020-bert-attack} introduce additional constraints to the perturbation space such as using synonyms for substitution, part-of-speech or semantic similarity checking to ensure the generated samples are semantically closed to the original ones.

Regarding the defenses against NLP attacks, adversarial training is one of the most successful defenses. The first adversarial training method is introduced in \citet{DBLP:journals/corr/GoodfellowSS14} for image data. The authors show that training on both clean and adversarial examples can improve the model's robustness. \citet{DBLP:conf/iclr/MiyatoDG17} develops a similar approach for the textual domain with $L_2$-bounded perturbation in the embedding space. \citet{jia-etal-2019-certified} and \citet{huang-etal-2019-achieving} propose using axis-aligned bounds to restrict adversarial perturbation. However, \citet{DBLP:conf/iclr/DongLJ021} later argues that these bounds are not sufficiently inclusive or exclusive. Therefore, the authors propose to instead model the perturbation space as the convex hull of word synonyms. They also propose an entropy-based regularization to encourage perturbations to point to actual valid words. \citet{DBLP:conf/iclr/ZhuCGSGL20} proposes FreeLB, a novel adversarial training approach that addresses the computational inefficiency of previous methods. However, their original work focuses on improving generalization rather than robustness.

\subsection{Off-manifold Conjecture and Manifold-based Defenses}
The off-manifold conjecture was first proposed in \citet{https://doi.org/10.48550/arxiv.1608.07690} as an alternative explanation for the existence of adversarial examples to previous ones \cite{DBLP:journals/corr/GoodfellowSS14, DBLP:journals/corr/SzegedyZSBEGF13}. Although it has been shown in previous works that on-manifold adversarial examples do exist \cite{DBLP:conf/iclr/GilmerMFSRWG18}, \citet{Stutz_2019_CVPR} later shows that on-manifold robustness is essentially generalization, i.e., they are simply generalization errors. Recent work from \citet{DBLP:journals/corr/abs-2106-10151} independently finds similar observations to \citet{https://doi.org/10.48550/arxiv.1608.07690} and \citet{Stutz_2019_CVPR}. They propose a conceptual framework called Dimpled Manifold Model and use it to explain various unanswered phenomena of adversarial examples \cite{DBLP:conf/nips/IlyasSTETM19, DBLP:conf/iclr/TsiprasSETM19}.

Based on this property, many defenses have been developed. \citet{DBLP:conf/iclr/SamangoueiKC18} proposes Defense-GAN, which uses a Wasserstein Generative Adversarial Network (WGAN) \cite{pmlr-v70-arjovsky17a} to project adversarial examples onto the approximated manifold. A similar idea is used in \citet{DBLP:conf/iclr/SchottRBB19}, in which the authors propose to use Variational Auto-Encoder (VAE) \cite{DBLP:journals/corr/KingmaW13} to learn the data manifold.
\section{Textual Manifold-based Defense}
\label{methodology}

In this section, we introduce our method in greater detail. We first introduce how we approximate the contextualized embedding manifold using deep generative models. Then, we describe how to perform on-manifold projection with the trained model in the inference phase. The general overview of our method is visualized in Figure \ref{fig:overview}. 

\subsection{Textual Manifold Approximation}
Approximating the manifold of textual data is not straightforward due to its discrete nature. Instead of modeling at the sentence level, we relax the problem by mapping texts into their continuous representations.

More formally, let $F: \mathcal{X} \rightarrow \mathbb{R}^n$ be a language model that maps an input text $x \in \mathcal{X}$ to its corresponding $n$-dimensional embedding $t$, where $\mathcal{X}$ denotes the set of all texts, we first compute the continuous representations $t$ for all $x$ in the dataset. We assume that all $t$ lie along a low-dimensional manifold $\mathcal{T} \subset \mathbb{R}^n$. Several studies have shown that the contextualized embedding space consists of disjoint clusters \cite{DBLP:conf/icml/MamouLRSTKC20,DBLP:conf/iclr/CaiHB021}. Based on these findings, we assume that $\mathcal{T}$ is a union of disconnected submanifolds, i.e., $\mathcal{T} = \bigcup \mathcal{T}_i,\; \forall i \ne j: \mathcal{T}_i\cap \mathcal{T}_j = \emptyset$. The problem is now simplified to approximating the contextualized embedding manifold $\mathcal{T}$. However, choosing the appropriate generative model is crucial as learning complex manifolds in high dimensional spaces heavily depends on the underlying geometry \cite{b2e6b962d8934edfb6b70d6112436f60}. While previous manifold-based defenses for images are able to achieve impressive results with simple models such as WGAN \cite{DBLP:conf/iclr/SamangoueiKC18} or VAE \cite{DBLP:conf/iclr/SchottRBB19}, we will show in Section \ref{the_importance_of_disconnectedness} that this is not the case for contextualized embeddings. 

\begin{proposition}
\label{prop:1}
Let $\mathcal{X}$ and $\mathcal{Y}$ be topological spaces and let $f: \mathcal{X} \rightarrow \mathcal{Y}$ be a continuous function. If $\mathcal{X}$ is connected then the image $f(\mathcal{X})$ is connected.
\end{proposition}

The original GAN proposed by \citet{NIPS2014_5ca3e9b1} trains a continuous generator $G: \mathcal{Z} \rightarrow \mathcal{X}'$ that maps a latent variable $z \in \mathcal{Z}$ sampled from some prior $P(z)$ to the target space $\mathcal{X}'$. The standard multivariate normal $\mathcal{N}(0, I)$ is often chosen as the distribution for $P(z)$. This implies that $P(z)$ is supported on the connected space $\mathbb{R}^d$. Therefore, the target space $\mathcal{X}'$ is also connected according to Proposition \ref{prop:1}. This explains why such simple model fails to learn disconnected manifolds  \cite{DBLP:conf/cvpr/GurumurthySB17,DBLP:conf/nips/KhayatkhoeiSE18,DBLP:conf/icml/TanielianIDM20}.

To approximate $\mathcal{T}$, we need to introduce disconnectedness in the latent space $\mathcal{Z}$. We follow the InfoGAN \cite{DBLP:conf/nips/ChenCDHSSA16} approach by adding an additional discrete latent code $c$ to $c \sim Cat(K, r)$, where $c$ is a $K$-dimensional one-hot vector, $K$ is a hyperparameter for the number of submanifolds in $\mathcal{T}$ and $r \in \mathbb{R}^K$ denotes the probabilities for each value of $c$ $\left (r_i \ge 0, \sum r_i = 1\right )$. The generator now becomes $G: \mathcal{Z} \times \mathcal{C} \rightarrow \mathcal{T}$.

Naturally, we expect to have different values of $c$ targeting different submanifold $\mathcal{T}_i$. This can be achieved by maximizing the mutual information $I(c; G(z, c))$ between the latent code $c$ and the generated embeddings $G(z, c)$. However, directly optimize $I(c, G(z, c))$ can be difficult, so we instead maximize the lower bound of $I(c, G(z, c))$:
\begin{equation}
\begin{split}
&I(c;G(z, c)) \\
&= H(c) - H(c\mid G(z, c)) \\
&= \mathbb{E}_{t\sim P_g(t)}\left [ \mathbb{E}_{c'\sim P(c\mid t)}\left [ \log{P(c'\mid t)} \right ] \right ]+H(c) \\
&= \mathbb{E}_{t\sim P_g(t)}\left [ D_{\text{KL}}\left ( P(\cdot \mid t)\parallel Q(\cdot \mid t)\right ) \right. \\
&\quad + \left. \mathbb{E}_{c'\sim P(c\mid t)}\left [ \log{Q(c'\mid t)} \right ]\right ] + H(c) \\
&\ge \mathbb{E}_{t\sim P_g(t)}\left [ \mathbb{E}_{c'\sim P(c\mid t)}\left [ \log{Q(c'\mid t)} \right ] \right ] + H(c)  \\
&= \mathbb{E}_{c\sim P(c), t\sim P_g(t)}\left [ \log(Q(c\mid t)) \right ] + H(c)
\end{split}
\end{equation}
where $Q(c\mid t)$ is an auxiliary distribution parameterized by a neural network to approximate $P(c\mid t)$, $P_g(t)$ denotes the distribution of generated embeddings from $G(z, c)$. The problem of maximizing $I(c, G(z, c))$ becomes maximizing the following information-theoretic regularization:
\begin{equation}
L_I(G, Q) =  \mathbb{E}_{c\sim P(c), t\sim P_g(t)}\left [ \log(Q(c\mid t)) \right ]
\end{equation}
Combine with the objective function of GAN,
\begin{equation}
\begin{split}
V(D,G)&= \mathbb{E}_{t\sim P_r(t)}\left [ \log{\left ( D\left ( t \right ) \right )} \right ] \\
&\quad + \mathbb{E}_{\hat{t}\sim P_g(t)}\left [ \log{\left ( 1-D\left ( \hat{t} \right ) \right )} \right ]
\end{split}
\end{equation}
where $P_r(t)$ denotes the distribution of natural text embeddings, we obtain the following initial objective function to implicitly learn the manifold $\mathcal{T}$:
\begin{equation}
\min_{G,Q}\max_D V(D, G) - \lambda L_I(G,Q)
\label{eqn:initial_obj}
\end{equation}

\subsection{Learnable Prior}
One limitation with the original InfoGAN design is the fixed uniform latent code distribution. Firstly, the true number of submanifolds in the target space is generally unknown. Secondly, data samples are not likely to be uniformly distributed over all submanifolds. To address these issues, we use expectation maximization to learn the optimal prior. Since $Q(c\mid t)$ approximates $P(c\mid t)$, $\mathbb{E}_{t\sim P_r}Q(c\mid t)$ gives us an approximation for the true prior distribution $P(c)$ which will then be used to train the InfoGAN in the M step. Doing this, the probabilities $r_i$ are adaptively readjusted and redundant latent code values will have their weights zeroed out. Instead of optimizing $r$ directly, we model it as $r = \text{softmax}(\hat{r}),\;\hat{r} \in \mathbb{R}^K$ then optimize on the unconstrained $\hat{r}$. We train $\hat{r}$ by minimizing the following cross entropy:
\begin{equation}
\begin{split}
H(P(c), r) &= -\mathbb{E}_{c\sim P(c)}[\log{r}] \\ 
 &= -\mathbb{E}_{t\sim P_r(t),c\sim P(c\mid t)}[\log{r}] \\ 
 &= \mathbb{E}_{t\sim P_r(t)}[H(P(c\mid t), r)] \\
 &\approx  \mathbb{E}_{t\sim P_r(t)}[H(Q(c\mid t), r)]
\end{split}
\end{equation}
We obtain the objective function for prior learning as follow:
\begin{equation}
L_P(\hat{r}) = \mathbb{E}_{t\sim P_r(t)}\left [ H(Q(c\mid t), r)\right ]
\label{eqn:prior_learning}
\end{equation}
Combine Equation \ref{eqn:prior_learning} with Equation \ref{eqn:initial_obj}, we arrive at the final objective function for manifold approximation:
\begin{equation}
\min_{G,Q,\hat{r}}\max_D V(D, G) - \lambda L_I(G,Q) + L_P(\hat{r})
\end{equation}

\subsection{On-manifold Projection}
Once $G$ is trained, the next step is to develop an on-manifold projection method. Given an input embedding $t$, projecting it onto the approximated manifold is essentially finding an embedding $\hat{t}$ on $G(z, c)$ that is the closest to $t$, i.e., solving $\min_{z, c} \left \| G(z, c) - t\right \|_2$. Conveniently, we can utilize the auxiliary network $Q(c\mid t)$ to determine the optimal value for $c$, which is the submanifold id that $t$ belongs to. The problem simplified to $\min_{z} \left \| G(z, c_t) - t\right \|_2$, where $c_t = Q(c\mid t)$. For the latent variable $z$, previous works in the GAN inversion literature often use optimization-based approaches to find the optimal $z$ \cite{DBLP:conf/iclr/SamangoueiKC18,DBLP:journals/tnn/CreswellB19a,DBLP:conf/iccv/AbdalQW19}. However, we will show in Section \ref{gradient_vs_sampling_reconstruction} that simply sampling $k$ candidate $z_i \sim P(z)$ and selecting the one that produces minimal reconstruction loss results in better robustness and faster inference speed. In summary, given an embedding $t$, we compute the reconstructed on-manifold embedding $\hat{t}$ as follow:
\begin{equation}
\begin{aligned}
& \hat{t} = G(z^*, c_t)\\
\textrm{where} \quad & c_t = Q(c\mid t)\\
& z^* = \underset{z\sim P(z)}{\arg\min} \left \| G(z, c_t) - t \right \|_2\\
\label{eqn:on_manifold_projection}
\end{aligned}
\end{equation}
The reconstructed embedding $\hat{t}$ is then fed into a classifier $C: \mathcal{T} \rightarrow \mathcal{Y}$ to produce the final predicted label $\hat{y} \in \mathcal{Y}$.

\section{Experiments}
\label{experiments}

\subsection{Experimental Setting}

\paragraph{Datasets}
We evaluate our method on three datasets: AG-News Corpus (AGNEWS) \cite{NIPS2015_250cf8b5}, Internet Movie Database (IMDB) \cite{maas-etal-2011-learning}, and Yelp Review Polarity (YELP) \cite{10.5555/2969239.2969312}. The AGNEWS dataset contains over 120000 samples, each belonging to one of the four labels: World, Sports, Business, Sci/Tech. The IMDB dataset contains 50000 data samples of movie reviews with binary labels for negative and positive sentiments. The YELP dataset contains nearly 600000 samples of highly polar Yelp reviews with binary labels. However, due to limitations in computing resources, we only use a subset of 63000 samples of the YELP dataset. In addition, we randomly sample 10\% of the training set for validation in all datasets.

\begin{table*}[!ht]
\small
\centering
\setlength{\tabcolsep}{0.15cm}
\renewcommand{\arraystretch}{1.1}
\resizebox{\textwidth}{!}{%
\begin{tabular}{c||c||cccc||cccc||cccc}
\hline
\hline
\multirow{2}{*}{Model}   & \multirow{2}{*}{Defense} & \multicolumn{4}{c||}{AGNEWS}                                                                                                   & \multicolumn{4}{c||}{IMDB}                                                                                                     & \multicolumn{4}{c}{YELP}                                                                                                     \\ \cline{3-14} 
                         &                          & \multicolumn{1}{c|}{CA}             & \multicolumn{1}{c|}{PW}            & \multicolumn{1}{c|}{TF}            & BA            & \multicolumn{1}{c|}{CA}             & \multicolumn{1}{c|}{PW}            & \multicolumn{1}{c|}{TF}            & BA            & \multicolumn{1}{c|}{CA}             & \multicolumn{1}{c|}{PW}            & \multicolumn{1}{c|}{TF}            & BA            \\ \hline
\multirow{5}{*}{BERT}    & Vanilla                  & \multicolumn{1}{c|}{{\ul 94.39}}    & \multicolumn{1}{c|}{39.2}          & \multicolumn{1}{c|}{27.9}          & 39.0          & \multicolumn{1}{c|}{92.15}          & \multicolumn{1}{c|}{6.5}           & \multicolumn{1}{c|}{2.3}           & 1.0           & \multicolumn{1}{c|}{{\ul 95.28}}    & \multicolumn{1}{c|}{13.7}          & \multicolumn{1}{c|}{10.5}          & 2.8           \\ 
                         & ASCC                     & \multicolumn{1}{c|}{91.57}          & \multicolumn{1}{c|}{32.8}          & \multicolumn{1}{c|}{31.4}          & 32.1          & \multicolumn{1}{c|}{88.48}          & \multicolumn{1}{c|}{15.1}          & \multicolumn{1}{c|}{12.4}          & 11.2          & \multicolumn{1}{c|}{91.46}          & \multicolumn{1}{c|}{19.4}          & \multicolumn{1}{c|}{15.7}          & 12.2          \\ 
                         & DNE                      & \multicolumn{1}{c|}{94.09}          & \multicolumn{1}{c|}{34.0}          & \multicolumn{1}{c|}{33.6}          & {\ul 52.3}    & \multicolumn{1}{c|}{89.97}          & \multicolumn{1}{c|}{25.7}          & \multicolumn{1}{c|}{23.0}          & 20.6          & \multicolumn{1}{c|}{93.97}          & \multicolumn{1}{c|}{{\ul 33.3}}    & \multicolumn{1}{c|}{{\ul 31.2}}    & \textbf{43.8} \\ 
                         & SAFER                    & \multicolumn{1}{c|}{\textbf{94.42}} & \multicolumn{1}{c|}{{\ul 39.3}}    & \multicolumn{1}{c|}{{\ul 35.5}}    & 42.3          & \multicolumn{1}{c|}{\textbf{92.26}} & \multicolumn{1}{c|}{\textbf{41.4}} & \multicolumn{1}{c|}{{\ul 39.1}}    & {\ul 30.7}    & \multicolumn{1}{c|}{\textbf{95.39}} & \multicolumn{1}{c|}{29.8}          & \multicolumn{1}{c|}{25.8}          & 23.7          \\ 
                         & TMD                      & \multicolumn{1}{c|}{94.29}          & \multicolumn{1}{c|}{\textbf{70.0}} & \multicolumn{1}{c|}{\textbf{50.0}} & \textbf{55.2} & \multicolumn{1}{c|}{{\ul 92.17}}    & \multicolumn{1}{c|}{{\ul 38.7}}    & \multicolumn{1}{c|}{\textbf{44.2}} & \textbf{33.7} & \multicolumn{1}{c|}{95.24}          & \multicolumn{1}{c|}{\textbf{36.8}} & \multicolumn{1}{c|}{\textbf{40.9}} & {\ul 28.6}    \\ \hline
\multirow{5}{*}{RoBERTa} & Vanilla                  & \multicolumn{1}{c|}{\textbf{95.04}} & \multicolumn{1}{c|}{44.1}          & \multicolumn{1}{c|}{34.5}          & 44.5          & \multicolumn{1}{c|}{93.24}          & \multicolumn{1}{c|}{4.4}           & \multicolumn{1}{c|}{1.0}           & 0.1           & \multicolumn{1}{c|}{{\ul 96.64}}    & \multicolumn{1}{c|}{37.0}          & \multicolumn{1}{c|}{20.1}          & 9.0           \\ 
                         & ASCC                     & \multicolumn{1}{c|}{92.62}          & \multicolumn{1}{c|}{48.1}          & \multicolumn{1}{c|}{41.0}          & 49.1          & \multicolumn{1}{c|}{92.62}          & \multicolumn{1}{c|}{23.1}          & \multicolumn{1}{c|}{13.5}          & 11.8          & \multicolumn{1}{c|}{95.42}          & \multicolumn{1}{c|}{15.0}          & \multicolumn{1}{c|}{8.6}           & 4.5           \\ 
                         & DNE                      & \multicolumn{1}{c|}{94.93}          & \multicolumn{1}{c|}{{\ul 58.0}}    & \multicolumn{1}{c|}{{\ul 46.5}}    & {\ul 54.5}    & \multicolumn{1}{c|}{\textbf{94.20}} & \multicolumn{1}{c|}{48.8}          & \multicolumn{1}{c|}{26.9}          & 16.0          & \multicolumn{1}{c|}{\textbf{96.76}} & \multicolumn{1}{c|}{64.4}          & \multicolumn{1}{c|}{64.0}          & 45.2          \\ 
                         & SAFER                    & \multicolumn{1}{c|}{94.58}          & \multicolumn{1}{c|}{51.3}          & \multicolumn{1}{c|}{41.9}          & 46.1          & \multicolumn{1}{c|}{{\ul 93.92}}    & \multicolumn{1}{c|}{{\ul 52.8}}    & \multicolumn{1}{c|}{{\ul 47.1}}    & {\ul 40.6}    & \multicolumn{1}{c|}{96.59}          & \multicolumn{1}{c|}{{\ul 65.6}}    & \multicolumn{1}{c|}{{\ul 67.9}}    & {\ul 48.3}    \\ 
                         & TMD                      & \multicolumn{1}{c|}{{\ul 95.03}}    & \multicolumn{1}{c|}{\textbf{68.3}} & \multicolumn{1}{c|}{\textbf{54.0}} & \textbf{56.7} & \multicolumn{1}{c|}{93.26}          & \multicolumn{1}{c|}{\textbf{60.5}} & \multicolumn{1}{c|}{\textbf{66.8}} & \textbf{51.6} & \multicolumn{1}{c|}{96.62}          & \multicolumn{1}{c|}{\textbf{68.9}} & \multicolumn{1}{c|}{\textbf{70.9}} & \textbf{51.0} \\ \hline
\multirow{5}{*}{XLNet}   & Vanilla                  & \multicolumn{1}{c|}{{\ul 94.80}}    & \multicolumn{1}{c|}{34.4}          & \multicolumn{1}{c|}{28.0}          & 37.9          & \multicolumn{1}{c|}{{\ul 93.59}}    & \multicolumn{1}{c|}{7.1}           & \multicolumn{1}{c|}{2.3}           & 1.4           & \multicolumn{1}{c|}{{\ul 96.23}}    & \multicolumn{1}{c|}{28.0}          & \multicolumn{1}{c|}{14.0}          & 7.2           \\ 
                         & ASCC                     & \multicolumn{1}{c|}{92.64}          & \multicolumn{1}{c|}{38.6}          & \multicolumn{1}{c|}{33.4}          & 41.6          & \multicolumn{1}{c|}{92.57}          & \multicolumn{1}{c|}{15.8}          & \multicolumn{1}{c|}{11.1}          & 10.5          & \multicolumn{1}{c|}{95.52}          & \multicolumn{1}{c|}{39.3}          & \multicolumn{1}{c|}{24.2}          & 12.8          \\ 
                         & DNE                      & \multicolumn{1}{c|}{\textbf{94.99}} & \multicolumn{1}{c|}{{\ul 48.8}}    & \multicolumn{1}{c|}{{\ul 38.4}}    & {\ul 44.1}    & \multicolumn{1}{c|}{93.53}          & \multicolumn{1}{c|}{\textbf{42.9}} & \multicolumn{1}{c|}{{\ul 33.2}}    & \textbf{26.6} & \multicolumn{1}{c|}{\textbf{96.64}} & \multicolumn{1}{c|}{{\ul 55.4}}    & \multicolumn{1}{c|}{{\ul 53.0}}    & {\ul 41.3}    \\ 
                         & SAFER                    & \multicolumn{1}{c|}{93.87}          & \multicolumn{1}{c|}{40.4}          & \multicolumn{1}{c|}{32.1}          & 38.1          & \multicolumn{1}{c|}{93.48}          & \multicolumn{1}{c|}{22.7}          & \multicolumn{1}{c|}{16.7}          & 6.7           & \multicolumn{1}{c|}{96.17}          & \multicolumn{1}{c|}{48.9}          & \multicolumn{1}{c|}{36.7}          & 23.7          \\ 
                         & TMD                      & \multicolumn{1}{c|}{94.57}          & \multicolumn{1}{c|}{\textbf{66.9}} & \multicolumn{1}{c|}{\textbf{54.2}} & \textbf{56.5} & \multicolumn{1}{c|}{\textbf{93.62}} & \multicolumn{1}{c|}{{\ul 25.3}}    & \multicolumn{1}{c|}{\textbf{37.9}} & {\ul 21.3}    & \multicolumn{1}{c|}{96.17}          & \multicolumn{1}{c|}{\textbf{61.4}} & \multicolumn{1}{c|}{\textbf{64.8}} & \textbf{51.2} \\ \hline
\hline
\end{tabular}%
}
\caption{The robustness of different defenses on AGNEWS, IMDB, and YELP. We denote the clean accuracy, accuracy under PWWS, TextFooler, Bert-Attack as CA, PW, TF, BA, respectively. The best performance for each model is \textbf{bolded}, and the second-best performance is \underline{underlined}.}
\label{tab:main_results}
\end{table*}

\paragraph{Model Architectures}
To test if our method is able to generalize to different architectures, we apply TMD on three state-of-the-art pretrained language models: $\text{BERT}_{\text{base}}$ \cite{devlin-etal-2019-bert}, $\text{RoBERTa}_{\text{base}}$ \cite{DBLP:journals/corr/abs-1907-11692}, and $\text{XLNet}_{\text{base}}$ \cite{NEURIPS2019_dc6a7e65}. BERT is a Transformer-based language model that has brought the NLP research by storm by breaking records on multiple benchmarks. Countless variants of BERT have been proposed \cite{xia-etal-2020-bert} in which RoBERTa and XLNet are two of the most well-known. In addition to the above models, we also experiment with their larger versions in Appendix \ref{additional_experiments}.

\paragraph{Adversarial Attacks}
We choose the following state-of-the-art attacks to measure the robustness of our method: (1) PWWS \cite{ren-etal-2019-generating} is a word synonym substitution attack where words in a sentence are greedily replaced based on their saliency and maximum word-swap effectiveness. (2) TextFooler \cite{Jin_Jin_Zhou_Szolovits_2020} utilize nearest neighbor search in the counter-fitting embeddings \cite{mrksic-etal-2016-counter} to construct the  dictionaries. Subsequently, words are swapped greedily based on their importance scores. They also introduce constraints such as part-of-speech and semantic similarity checking to ensure the generated adversarial example looks natural and does not alter its original label. (3) BERT-Attack \cite{li-etal-2020-bert-attack} is similar to TextFooler, but instead of using synonyms for substitution, they use BERT masked language model to produce candidate words that fit a given context. They also introduce subtle modifications to the word importance scoring function. All attacks above are implemented using the TextAttack framework \cite{morris-etal-2020-textattack}. 

For a fair comparison, we follow the same settings as \citet{li-etal-2021-searching}, in which all attacks must follow the following constraints: (1) the maximum percentage of modified words $\rho_{max}$ for AGNEWS, IMDB, and YELP must be $0.3$, $0.1$, and $0.1$ respectively, (2) the maximum number of candidate replacement words $K_{max}$ is set to $50$, (3) the minimum semantic similarity\footnote{The semantic similarity between $x$ and $x'$ is approximated by measuring the cosine similarity between their text embeddings produced from Universal Sentence Encoder \cite{cer-etal-2018-universal}} $\epsilon_{min}$ between original input $x$ and adversarial example $x'$ must be $0.84$, and (4) the maximum number of queries to the victim model is $Q_{max} = K_{max} \times L$, where $L$ is the length of the original input $x$.

\paragraph{Baseline Defenses}
We compare our method with other families of defenses. For adversarial training defenses, we choose the Adversarial Sparse Convex Combination (ASCC) \cite{DBLP:conf/iclr/DongLJ021} and Dirichlet Neighborhood Ensemble (DNE) \cite{zhou-etal-2021-defense}. Both methods model the perturbation space as the convex hull of word synonyms. The former introduces an entropy-based sparsity regularizer to better capture the geometry of real word substitutions. The latter expands the convex hull for a word $x$ to cover all synonyms of $x$'s synonyms and combines Dirichlet sampling in this perturbation space with adversarial training to improve the model robustness. For certified defenses, we choose SAFER \cite{ye-etal-2020-safer}, which is a method based on the randomized smoothing technique. Given an input sentence, SAFER constructs a set of randomized inputs by applying random synonym substitutions and leveraging statistical properties of the predicted labels to certify the robustness.

\begin{figure*} [ht] 
    \centering
    \captionsetup[subfigure]{oneside,margin={0.9cm,0cm}}
    \begin{subfigure}{0.32\textwidth}
        \centering
        \includegraphics[height=3.7cm]{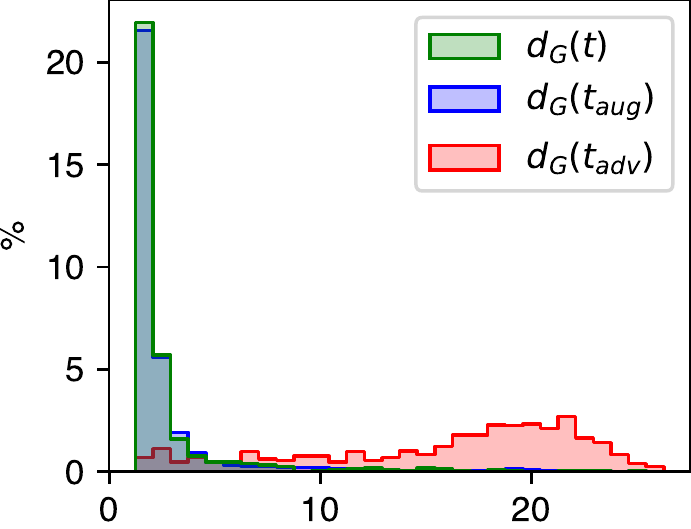}
        \caption{BERT}
        \label{fig:imdb_bert_bertattack}
    \end{subfigure}
    \captionsetup[subfigure]{oneside,margin={0.65cm,0cm}}
    \hspace{0.1cm}
    \begin{subfigure}{0.32\textwidth}
        \centering
        \includegraphics[height=3.7cm]{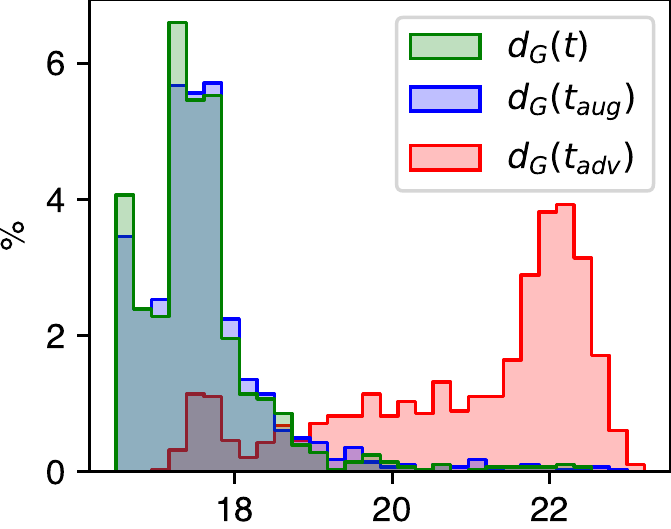}
        \caption{RoBERTa}
        \label{fig:imdb_roberta_bertattack}
    \end{subfigure}
    \captionsetup[subfigure]{oneside,margin={1.2cm,0cm}}
    \hspace{0.1cm}
    \begin{subfigure}{0.32\textwidth}
        \centering
        \includegraphics[height=3.7cm]{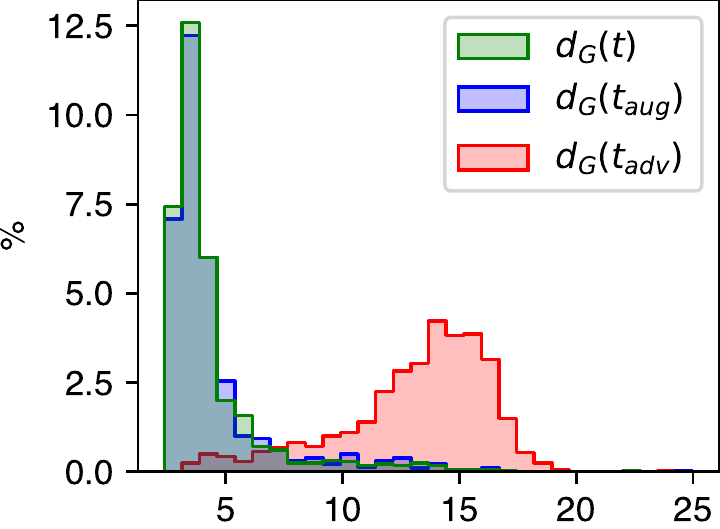}
        \caption{XLNet}
        \label{fig:imdb_xlnet_bertattack}
    \end{subfigure}
    \hspace{0.35cm}
    \caption{The distributions of distance-to-manifold of clean, augmented, and adversarial embeddings in the IMDB dataset. The adversarial examples are generated using BERT-Attack \cite{li-etal-2020-bert-attack}. Augmented examples are generated using random synonym substitution.}
    \label{fig:manifold_assumption}
\end{figure*}
\paragraph{Implementation Details}
In practice, we pre-compute the text embeddings for all inputs to reduce the computational cost. For BERT and RoBERTa, we use the [CLS] representation as the text embedding. For XLNet, we use the last token's representation. More implementation details such as training InfoGAN, choosing hyperparameters can be found in Appendix \ref{implementation_details}.

\subsection{Main Results}
To evaluate the robustness of different defenses, we randomly select 1000 samples from the test set and evaluate their accuracy under attacks. For the clean accuracy, we evaluate it on the entire test set.

The results for the AGNEWS dataset are shown in Table \ref{tab:main_results}. We denote the "Vanilla" method as the original model without any defense mechanism. As we can see from the results, TMD outperforms previous methods under various settings by a large margin. Despite a slight decline in the clean accuracy, TMD achieves state-of-the-art robustness for BERT, RoBERTa, and XLNet with 23.03\%, 18.63\%, and 25.77\% average performance gain over all attacks, respectively.

Interestingly, slightly different trends are found in the IMDB and YELP datasets. First of all, all models are generally more vulnerable to adversarial examples. This could be explained by the long average sentence length in IMDB (313.87 words) and YELP (179.18 words). This value is much larger than the AGNEWS, about 53.17 words. Longer sentences result in less restricted perturbation space for the attacker to perform word-substitution attacks, hence increasing the attack success rate. Regarding robustness, our method outperforms other methods in the majority of cases.

\subsection{Testing the Off-Manifold Conjecture in NLP}

Equipped with a way to approximate the contextualized embedding manifold, we aim to validate the off-manifold conjecture in the NLP domain. To achieve this, we first define the distance of an embedding $t$ to the approximated manifold $G(z, c)$ as 

\begin{equation}
d_G(t) = \left \| t-\hat{t} \right \|_2
\label{eqn:distance_to_manifold}
\end{equation}

where $\hat{t}$ is the on-manifold projection computed from Equation \ref{eqn:on_manifold_projection}. The off-manifold conjecture states that adversarial examples tend to leave the underlying natural data manifold. Therefore, if the off-manifold conjecture holds, we should observe small $d_G(t)$ for clean examples, while adversarial examples $t_{adv}$ have large values of $d_G(t_{adv})$. 

For each sentence $x$ in the test set, we find its corresponding adversarial example $x_{adv}$. Additionally, to ensure that large $d_G(t_{adv})$ is not simply caused by word substitution, we randomly substitute $x$ with synonyms to make an augmented sentence $x_{aug}$ and see if the resulted distribution $d_G(t_{aug})$ diverges away from $d_G(t)$. We set the modifying ratio equal to the average percentage of perturbed words in $x_{adv}$ for a fair comparison. The distributions of $d_G(t)$, $d_G(t_{aug})$, and $d_G(t_{adv})$ are visualized in Figure \ref{fig:manifold_assumption}. 

From the figure, we can see a clear divergence between the distribution of $d_G(t)$ and $d_G(t_{adv})$ on all models. Furthermore, the distribution $d_G(t_{aug})$ remains nearly identical to $d_G(t)$. This shows that simple word substitution does not cause the embedding to diverge off the natural manifold. Additionally, it is important to emphasize that since all $x$ are unseen examples from the test set, low values of $d_G(t)$ are not simply due to overfitting in the generative model. These results provide empirical evidence to support the off-manifold conjecture in NLP.

\subsection{The Importance of Disconnectedness in Contextualized Manifold Approximation}
\label{the_importance_of_disconnectedness}

\begin{table}[t]
    \small
    \centering
    \setlength{\tabcolsep}{0.2cm}
    \renewcommand{\arraystretch}{1.2}
    \begin{tabular}{cc|c|c|c}
        \hline
        \hline
        \multicolumn{2}{c|}{Method}                                 & RL              & CLN            & AUA                \\ 
        \hline
        \multicolumn{1}{c|}{\multirow{2}{*}{BERT}}    & TMD-InfoGAN & \textbf{3.121}  & \textbf{92.15} & \textbf{33.70} \\
        \multicolumn{1}{c|}{}                         & TMD-DCGAN   & 3.707           & 92.04          & 4.30           \\ 
        \hline
        \multicolumn{1}{c|}{\multirow{2}{*}{RoBERTa}} & TMD-InfoGAN & \textbf{17.678} & \textbf{93.24} & \textbf{51.60} \\
        \multicolumn{1}{c|}{}                         & TMD-DCGAN   & 19.644          & 93.07          & 2.90           \\ 
        \hline
        \multicolumn{1}{c|}{\multirow{2}{*}{XLNet}}   & TMD-InfoGAN & \textbf{4.679}  & \textbf{93.59} & \textbf{19.20} \\
        \multicolumn{1}{c|}{}                         & TMD-DCGAN   & 7.056           & 93.47          & 4.00           \\ 
        \hline
        \hline
    \end{tabular}
    \caption{The accuracy under attack comparison between TMD-InfoGAN with disconnected support and TMD-DCGAN with connected support under BERT-Attack. We denote the reconstruction loss, clean accuracy, and accuracy under attack as RL, CLN, and AUA, respectively. All results are measured on the IMDB dataset.}
    \label{tab:infogan_vs_dcgan}
\end{table}

In this experiment, we study the significance of using disconnected generative models in TMD to improve robustness. We replace InfoGAN with DCGAN \cite{DBLP:journals/corr/RadfordMC15}, which has a Gaussian prior distribution $z \sim \mathcal{N}(0, I)$ supported on the connected space $\mathbb{R}^d$ and has the same backbone as InfoGAN for a fair comparison. We then measure its reconstruction capability, the resulted generalization, robustness and compare the differences between the two versions of TMD. The reconstruction loss is defined similarly as Equation \ref{eqn:distance_to_manifold}.

As shown in Table \ref{tab:infogan_vs_dcgan}, DCGAN performs worse than InfoGAN on approximating the contextualized embedding manifold of all language models, which leads to a degradation in clean accuracy and a significant drop in robustness against adversarial attacks. The robustness of BERT, RoBERTa, and XLNET under TMD-DCGAN drop by 87.24\%, 94.38\%, and 79.17\%, respectively. This result is consistent with previous works on approximating disconnected manifold \cite{DBLP:conf/nips/KhayatkhoeiSE18, DBLP:conf/icml/TanielianIDM20} and shows that using disconnected generative models is crucial for robustness improvement in TMD.

\subsection{The Effect of Sampling Size $k$ on Robustness}

\begin{figure}[t]
    \centering
    \includegraphics[width=0.9\columnwidth]{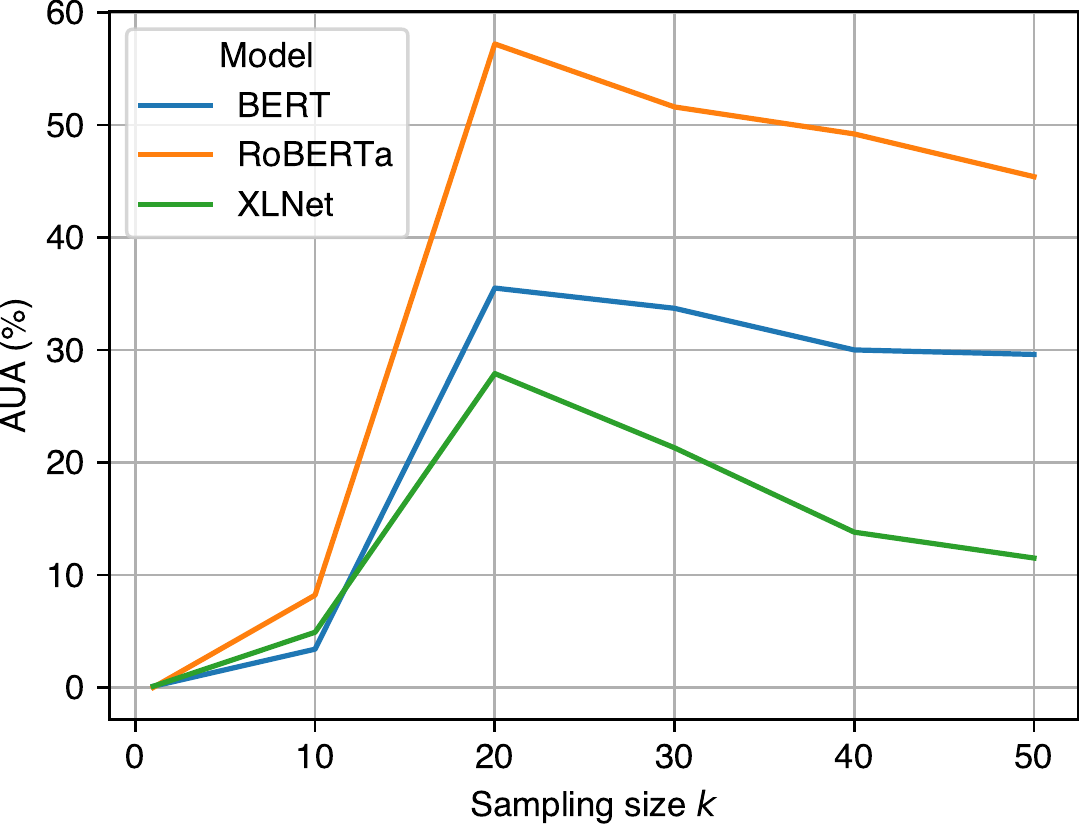}
    \caption{The relationship between the sampling size $k$ and the accuracy under BERT-Attack for different language models.}
    \label{fig:k_effect}
\end{figure}

We now study the effect of different $k$ values on TMD performance. From Figure \ref{fig:k_effect}, we can see an optimal value for $k$ across all models. A larger sampling size does not correspond with better robustness but slightly degrades it. We hypothesize that when $k$ is set too large, $z$ may be sampled from low-probability regions in the latent space, producing smaller reconstruction loss but not necessarily lying on the true embedding manifold. On the other hand, when $k$ is too small, it affects the reconstruction quality of InfoGAN, and the projected embedding may not well reassemble the original one.

\subsection{Comparison with Optimization-based Reconstruction Strategy}
\label{gradient_vs_sampling_reconstruction}
\begin{table}[ht]
    \small
    \centering
    \setlength{\tabcolsep}{0.2cm}
    \renewcommand{\arraystretch}{1.2}
    \begin{tabular}{c|c}
    \hline
    \hline
    Method              & BERT-Attack    \\ 
    \hline
    TMD                 & 33.70          \\ 
    \hline
    TMD-GD ($\alpha=1, N=10$)     & 26.50          \\ 
    TMD-GD ($\alpha=0.1, N=10$)   & 33.10          \\ 
    TMD-GD ($\alpha=0.01, N=10$)  & 30.60          \\ 
    TMD-GD ($\alpha=0.001, N=10$) & 31.20          \\ 
    \hline
    \hline
    \end{tabular}
    \caption{Comparison between sampling-based and optimization-based reconstruction strategies}
    \label{tab:different_reconstructions}
\end{table}

In addition to the sampling-based reconstruction, we also experiment with the optimization-based approach, one of the most common methods in the GAN inversion literature \cite{DBLP:journals/corr/abs-2101-05278,DBLP:journals/tnn/CreswellB19a,DBLP:conf/iccv/AbdalQW19}. Particularly, we evaluate the effectiveness of a reconstruction method similar to DefenseGAN \citet{DBLP:conf/iclr/SamangoueiKC18}. Given an embedding $t$, we sample $k$ initial values for $z$ from the prior distribution. We then perform $N$ steps of gradient descent (GD) with step size $\alpha$ separately on each $k$ value of $z$ to minimize the reconstruction loss. The optimal latent variable z is chosen to compute the final on-manifold projection $\hat{t}$. To put it shortly, this reconstruction method extends our method in Equation \ref{eqn:on_manifold_projection} with an additional step of GD optimization. We refer to this version of TMD as TMD-GD and compare several hyperparameter settings of this optimization-based approach with our sampling-based approach.

As can be seen in Table \ref{tab:different_reconstructions}, the TMD-GD reconstruction is outperformed by our sampling-based approach. We hypothesize that since TMD-GD does not consider the probability over the latent space, it can move the latent variables to low-probability regions, producing embeddings that may not lie on the true natural manifold. This problem is similar to sampling-based reconstruction with too large sampling size. Another drawback of TMD-GD is the additional computational cost introduced by GD, which can negatively affect the inference phase.
\section{Conclusion}
\label{conclusion}
In this paper, we show empirical evidence that the manifold assumption indeed holds for the textual domain. Based on this observation, we propose a novel method that projects input embeddings onto the approximated natural embedding manifold before classification to defend against adversarial examples. Extensive experiments show that our method consistently outperforms previous defenses by a large margin under various adversarial settings. Future research on developing better manifold approximation and on-manifold projection methods are interesting directions to further improve the robustness of this type of defense. We hope that the findings in this work can enable broader impacts on improving NLP robustness.

\section*{Limitations}
Despite the many advantages of TMD, it still has some limitations that can be improved. One problem, in particular, is to reduce the computational overhead of on-manifold projection. Since we are adding an additional reconstruction step, it adds latency to the inference phase. Other interesting questions that have not been fully addressed in this work due to time constraints include the effect of TMD when applying reconstruction to intermediate layers, alternative methods to construct text embeddings (e.g., by averaging all token embeddings instead of using [CLS] token), more sophisticated choices of manifold approximation models and reconstruction methods. These are interesting research directions that can extend the understanding and effectiveness of TMD.

\section*{Acknowledgements}
This work is supported by the Singapore Ministry of Education (MOE) Academic Research Fund
(AcRF) Tier 1 (RS21/20).

\bibliography{anthology,custom}
\bibliographystyle{acl_natbib}

\appendix

\section{Implementation Details}
\label{implementation_details}

\subsection{Preparing Datasets}
We use off-the-shelf datasets from HuggingFace Datasets \cite{lhoest-etal-2021-datasets}. Based on the average text length of each dataset, we set the model's max length to 128, 256, and 256 for AGNEWS, IMDB, and YELP, respectively.

\subsection{Finetuning Language Models to Downstream Tasks}
For pretrained language models, we utilize the base models provided by HuggingFace Transformers \cite{wolf-etal-2020-transformers}. We then finetune them on downstream tasks for ten epochs. The optimal learning rates for each pair of datasets and model are achieved using a simple grid search from 1e-6 to 9e-4. The optimally finetuned models are kept for robustness evaluation.

\subsection{Training InfoGAN}

\begin{table}[ht]
\small
\centering
\begin{tabular}{c|l}
\hline
\hline
Hyperparameter            & Values                                                                                   \\ \hline
$\alpha_g$         & \begin{tabular}[c]{@{}l@{}}1e-5, 3e-5, 5e-5, 7e-5,\\ 9e-5, 1e-4, 3e-4, 5e-4\end{tabular} \\ \hline
$\alpha_d$         & \begin{tabular}[c]{@{}l@{}}7e-5, 9e-5, 1e-4, 3e-4,\\ 5e-4, 7e-4, 9e-4, 1e-3\end{tabular} \\ \hline
$\alpha_p$     & 0, 1e-4, 1e-3, 1e-2                                                                      \\ \hline
$K$ & 50, 100, 200                                                                             \\ \hline
$d$     & 10, 20, 30, 40                                                                           \\ \hline
$d/g$            & 1, 2, 3, 4, 5                                                                            \\ \hline
\hline
\end{tabular}
\caption{The value ranges for random search on hyperparameters, where $\alpha_g$, $\alpha_d$, $\alpha_p$, $K$, $d$, $d/g$ denote the $G$'s learning rate, $D$'s learning rate, Prior's learning rate, latent variable $z$ dimension, latent code $c$ dimension, and the number of $D$'s iterations per $G$ updates.}
\label{tab:hparams_search}
\end{table}

Since InfoGAN uses the DCGAN backbone, we find it relatively stable during training. The only additional training trick we employ is tuning the ratio of discriminator iterations per generator update. We perform a random search on the hyperparameters, where their value ranges are shown in Table \ref{tab:hparams_search}. Additionally, the number of training epochs is set to 100 for all experiments. In general, we find that the most significant hyperparameters are $\alpha_g$, $\alpha_d$, and $d/g$, while $\alpha_p$, $K$, and $d$ do not contribute too much to the final robustness. Any reasonably large value for $d$ dimension is sufficient since redundant $c$ values eventually vanished during prior learning. The optimal hyperparameters after random search are shown in Table \ref{tab:hparams}. The detailed architectures for $G$ and $D$ are shown in Tables \ref{gen_arch} and \ref{dis_arch}.

\section{Additional Experiments}
\label{additional_experiments}

\subsection{Runtime Analysis}
\begin{table}[ht]
    \small
    \centering
    \setlength{\tabcolsep}{0.2cm}
    \renewcommand{\arraystretch}{1.2}
    \begin{tabular}{c|c}
    \hline
    \hline
    Defense              & Runtime (s)    \\ 
    \hline
    Vanilla                 & 27.21          \\ 
    ASCC                 & 460.76          \\ 
    SAFER                 & 27.07          \\ 
    DNE                 & 59.77          \\ 
    TMD                 & 56.19          \\ 
    \hline
    \hline
    \end{tabular}
    \caption{Inference speed comparison with other defenses. Tested with BERT on 1000 samples from the IMDB dataset.}
    \label{tab:inference_speed}
\end{table}
In this experiment, we want to measure how much overhead TMD introduces to the inference phase compared to other defenses. We sample 1000 inputs from the IMDB dataset and record the inference speed of BERT when equipped with different defenses. The results are shown in Table \ref{tab:inference_speed}. Despite an additional latency in the inference phase, TMD still achieves competitive performance.

\begin{table*}[h]
\centering
\resizebox{\textwidth}{!}{%
\begin{tabular}{c|cccccc|cccccc|cccccc}
\hline
\hline
\multicolumn{1}{l|}{\multirow{2}{*}{}} & \multicolumn{6}{c|}{AGNEWS}                                                                                                                                       & \multicolumn{6}{c|}{IMDB}                                                                                                                                         & \multicolumn{6}{c}{YELP}                                                                                                                                         \\ \cline{2-19} 
\multicolumn{1}{l|}{}                  & \multicolumn{1}{c|}{$\alpha_g$} & \multicolumn{1}{c|}{$\alpha_d$} & \multicolumn{1}{c|}{$\alpha_p$} & \multicolumn{1}{c|}{$d/g$} & \multicolumn{1}{c|}{$K$} & $d$ & \multicolumn{1}{c|}{$\alpha_g$} & \multicolumn{1}{c|}{$\alpha_d$} & \multicolumn{1}{c|}{$\alpha_p$} & \multicolumn{1}{c|}{$d/g$} & \multicolumn{1}{c|}{$K$} & $d$ & \multicolumn{1}{c|}{$\alpha_g$} & \multicolumn{1}{c|}{$\alpha_d$} & \multicolumn{1}{c|}{$\alpha_p$} & \multicolumn{1}{c|}{$d/g$} & \multicolumn{1}{c|}{$K$} & $d$ \\ \hline
BERT                                    & \multicolumn{1}{c|}{1e-4}       & \multicolumn{1}{c|}{1e-4}       & \multicolumn{1}{c|}{0}          & \multicolumn{1}{c|}{1}     & \multicolumn{1}{c|}{50}  & 20  & \multicolumn{1}{c|}{1e-4}       & \multicolumn{1}{c|}{1e-4}       & \multicolumn{1}{c|}{0}          & \multicolumn{1}{c|}{1}     & \multicolumn{1}{c|}{50}  & 40  & \multicolumn{1}{c|}{1e-4}       & \multicolumn{1}{c|}{9e-5}       & \multicolumn{1}{c|}{0}          & \multicolumn{1}{c|}{1}     & \multicolumn{1}{c|}{100} & 20  \\ 
RoBERTa                                 & \multicolumn{1}{c|}{2e-4}       & \multicolumn{1}{c|}{2e-4}       & \multicolumn{1}{c|}{1e-2}       & \multicolumn{1}{c|}{1}     & \multicolumn{1}{c|}{100} & 20  & \multicolumn{1}{c|}{3e-4}       & \multicolumn{1}{c|}{3e-4}       & \multicolumn{1}{c|}{0}          & \multicolumn{1}{c|}{1}     & \multicolumn{1}{c|}{200} & 20  & \multicolumn{1}{c|}{3e-5}       & \multicolumn{1}{c|}{1e-3}       & \multicolumn{1}{c|}{1e-4}       & \multicolumn{1}{c|}{2}     & \multicolumn{1}{c|}{100} & 20  \\ 
XLNet                                   & \multicolumn{1}{c|}{1e-4}       & \multicolumn{1}{c|}{1e-4}       & \multicolumn{1}{c|}{1e-4}       & \multicolumn{1}{c|}{1}     & \multicolumn{1}{c|}{100} & 20  & \multicolumn{1}{c|}{1e-4}       & \multicolumn{1}{c|}{1e-4}       & \multicolumn{1}{c|}{1e-4}       & \multicolumn{1}{c|}{1}     & \multicolumn{1}{c|}{100} & 20  & \multicolumn{1}{c|}{5e-4}       & \multicolumn{1}{c|}{9e-5}       & \multicolumn{1}{c|}{1e-3}       & \multicolumn{1}{c|}{5}     & \multicolumn{1}{c|}{100} & 30  \\ \hline
\hline
\end{tabular}%
}
\caption{Optimal hyperparameters after random search.}
\label{tab:hparams}
\end{table*}

\begin{table*}[!ht]
\small
\centering
\vskip 0.15in
\setlength{\tabcolsep}{0.15cm}
\renewcommand{\arraystretch}{1.2}
\begin{tabular}{c|c|c|c|c|c|c|c}
\hline
\hline
Operation       & Kernel  & Strides & Padding & Feature Maps & Batch Norm. & Activation & Shared? \\ \hline
$G(z, c): z \sim P(z), c \sim P(c)$         &         &         &         &   $(K+d) \times 1$            &            &            &         \\ \hline
ConvTranspose1d & 32 $\times$ 32 & 1 $\times$ 1   & 0       & 512 $\times$ 32     & Y           & ReLU       & N       \\ 
ConvTranspose1d & 4 $\times$ 4   & 2 $\times$ 2   & 1       & 384 $\times$ 64     & Y           & ReLU       & N       \\ 
ConvTranspose1d & 4 $\times$ 4   & 2 $\times$ 2   & 1       & 256 $\times$ 128    & Y           & ReLU       & N       \\ 
ConvTranspose1d & 4 $\times$ 4   & 2 $\times$ 2   & 1       & 128 $\times$ 256    & Y           & ReLU       & N       \\ \hline
ConvTranspose1d & 5 $\times$ 5   & 3 $\times$ 3   & 1       & 1 $\times$ 768      & N           & Tanh       & N       \\ 
\hline
\hline
\end{tabular}
\caption{Model architecture of the Generator Network $G$}
\label{gen_arch}
\end{table*}

\begin{table*}[!ht]
\small
\centering
\vskip 0.15in
\setlength{\tabcolsep}{0.15cm}
\renewcommand{\arraystretch}{1.2}
\begin{tabular}{c|c|c|c|c|c|c|c}
\hline
\hline
Operation         & Kernel  & Strides & Padding & Feature Maps & Batch Norm. & Activation              & Shared? \\ \hline
$D(t), Q(t)$        &         &         &         & 1 $\times$ 768              &             &                         &         \\ \hline
Conv1d            & 5 $\times$ 5   & 3 $\times$ 3   & 1       & 128 $\times$ 256    & N           & LeakyReLU (slope = 0.2) & Y       \\
Conv1d            & 4 $\times$ 4   & 2 $\times$ 2   & 1       & 256 $\times$ 128    & Y           & LeakyReLU (slope = 0.2) & Y       \\
Conv1d            & 4 $\times$ 4   & 2 $\times$ 2   & 1       & 384 $\times$ 64     & Y           & LeakyReLU (slope = 0.2) & Y       \\
Conv1d            & 4 $\times$ 4   & 2 $\times$ 2   & 1       & 512 $\times$ 32     & Y           & LeakyReLU (slope = 0.2) & Y       \\
Conv1d            & 4 $\times$ 4   & 2 $\times$ 2   & 1       & 768 $\times$ 16     & Y           & LeakyReLU (slope = 0.2) & Y       \\ \hline
$D$ Conv1d          & 16 $\times$ 16 & 1 $\times$ 1   & 0       & 1 $\times$ 1        & N           & Sigmoid                 & N       \\
$Q$ Fully Connected &         &         &         &   $1 \times K$           &          N   & Softmax                 & N       \\
\hline
\hline
\end{tabular}
\caption{Model architecture of the Discriminator $D$ and Auxiliary Network $Q$}
\label{dis_arch}
\end{table*}

\subsection{TMD on Varying Model Sizes}
\begin{table*}[h]
    \small
    \centering
    \setlength{\tabcolsep}{0.11cm}
    \renewcommand{\arraystretch}{1.1}
    \begin{subtable}[h]{0.45\textwidth}
        \centering
        \begin{tabular}{c|c|c|c|c|c}
            \hline
            \hline
            Model                          & Defense & CA             & PW             & TF             & BA             \\ \hline
            \multirow{2}{*}{BERT-base}     & Vanilla & 92.15          & 6.50           & 2.30           & 1.00           \\ 
                                          & TMD     & \textbf{92.17} & \textbf{38.70} & \textbf{44.20} & \textbf{33.70} \\ \hline
            \multirow{2}{*}{BERT-large}    & Vanilla & 93.04          & 29.30          & 21.10          & 16.50          \\ 
                                          & TMD     & \textbf{93.14} & \textbf{50.10} & \textbf{58.20} & \textbf{44.10} \\ \hline
            \multirow{2}{*}{RoBERTa-base}  & Vanilla & 93.24          & 4.40           & 1.00           & 0.10           \\ 
                                          & TMD     & \textbf{93.26} & \textbf{60.50} & \textbf{66.80} & \textbf{51.60} \\ \hline
            \multirow{2}{*}{RoBERTa-large} & Vanilla & \textbf{95.05} & 31.73          & 12.01          & 4.10           \\ 
                                          & TMD     & 94.95          & \textbf{70.60} & \textbf{74.77} & \textbf{61.60} \\
            \hline
            \hline
        \end{tabular}
        \caption{IMDB}
        \label{tab:imdb_large_lm}
    \end{subtable}
    \hspace{0.5cm}
    \begin{subtable}{0.45\textwidth}
        \centering
        \begin{tabular}{c|c|c|c|c|c}
            \hline
            \hline
            Model                          & Defense & CA             & PW             & TF             & BA             \\ \hline
            \multirow{2}{*}{BERT-base}     & Vanilla & \textbf{94.39} & 39.20          & 27.90          & 39.00          \\ 
                                          & TMD     & 94.29          & \textbf{70.00} & \textbf{50.00} & \textbf{55.20} \\ \hline
            \multirow{2}{*}{BERT-large}    & Vanilla & \textbf{94.59} & 30.30          & 20.80          & 26.20          \\ 
                                          & TMD     & 92.83          & \textbf{56.40} & \textbf{30.50} & \textbf{30.00} \\ \hline
            \multirow{2}{*}{RoBERTa-base}  & Vanilla & \textbf{95.04} & 44.10          & 34.50          & 44.50          \\ 
                                          & TMD     & 95.03          & \textbf{68.30} & \textbf{54.00} & \textbf{56.70} \\ \hline
            \multirow{2}{*}{RoBERTa-large} & Vanilla & \textbf{95.34} & 52.40          & 35.40          & 40.70          \\ 
                                          & TMD     & 94.88          & \textbf{71.80} & \textbf{50.80} & \textbf{51.20} \\
            \hline
            \hline
        \end{tabular}
        \caption{AGNEWS}
        \label{tab:agnews_large_lm}
    \end{subtable}
    \caption{Performance on different architectural sizes.}
    \label{tab:large_lm}
\end{table*}

In this experiment, we want to support the significance of TMD by showing its generalization across different model architectural sizes. Since larger versions of language models often use a hidden size of 1024, we have to employ a larger version of InfoGAN to adopt the larger embeddings. The results are shown in Table \ref{tab:large_lm}. As we can see from the table, combining TMD with larger models results in impressive robustness against various attacks.

\end{document}